\documentclass{article}
\usepackage{multirow}
\usepackage{microtype}
\usepackage{graphicx}
\usepackage{subfigure}
\usepackage{booktabs} 
\usepackage{tikz}

\usetikzlibrary{arrows.meta, bending}
\usetikzlibrary{decorations.markings} 
\usepackage{hyperref}

\usepackage[accepted]{icml2024}

\usepackage{amsmath}
\usepackage{amssymb}
\usepackage{mathtools}
\usepackage{amsthm}
\usepackage{dsfont}
\usepackage{mathrsfs}

\usepackage[capitalize,noabbrev]{cleveref}

\theoremstyle{plain}

\theoremstyle{definition}

\theoremstyle{remark}

\usepackage[textsize=tiny]{todonotes}

\usepackage{multirow}
\newcommand{\tabincell}[2]{\begin{tabular}{@{}#1@{}}#2\end{tabular}}

\icmltitlerunning{On Size and Hardness Generalization in Unsupervised Learning for the Travelling Salesman Problem}

\begin{document}

\twocolumn[
\icmltitle{On Size and Hardness Generalization in Unsupervised Learning for the Travelling Salesman Problem}




\begin{icmlauthorlist}
\icmlauthor{Yimemg Min}{cornell}
\icmlauthor{Carla P. Gomes}{cornell}
\end{icmlauthorlist}

\icmlaffiliation{cornell}{Department of Computer Science, Cornell University, Ithaca 14850, USA}

\icmlcorrespondingauthor{Yimeng Min}{min@cs.cornell.edu}

\icmlkeywords{Travelling Salesman Problem, Graph Neural Networks, Combinatorial Optimization, Unsupervised Learning}

\vskip 0.3in
]



\printAffiliationsAndNotice{}  

\begin{abstract}
We study the generalization capability of Unsupervised Learning in solving the Travelling Salesman Problem (TSP). We use a Graph Neural Network (GNN) trained with a surrogate loss function to generate an embedding for each node. We use these embeddings to construct a heat map that indicates the likelihood of each edge being part of the optimal route. We then apply local search to generate our final predictions. Our investigation explores how different training instance sizes, embedding dimensions, and distributions influence the outcomes of Unsupervised Learning methods. Our results show that training with larger instance sizes and increasing embedding dimensions can build a more effective representation, enhancing the model's ability to solve TSP. Furthermore, in evaluating generalization across different distributions, we first determine the hardness of various distributions and explore how different hardnesses affect the final results. Our findings suggest that models trained on harder instances exhibit better generalization capabilities, highlighting the importance of selecting appropriate training instances in solving TSP using Unsupervised Learning.
\end{abstract}
\section{Introduction}
The goal of machine learning for Combinatorial Optimization (CO) is to enhance or surpass handcrafted heuristics.
Recently, there has been an increasing trend in applying Machine Learning (ML) to tackle CO problems~\cite{bengio2021machine}. Different from manually crafted heuristics, machine learning approaches harness the power of data to uncover patterns in CO problems.

The Euclidean Travelling Salesman Problem (TSP) is one of the most famous and intensively studied CO problems.  TSP asks the following question:
\emph{Given a list of cities and the distances between each pair of cities, what is the shortest possible route that visits each city exactly once and returns to the origin city?}
A variety of methods have been developed to solve TSP, including the Lin-Kernighan-Helsgaun (LKH) heuristics, which is known for their effectiveness in approximating solutions~\cite{helsgaun2000effective}, and the Concorde solver, which guarantees optimality of the solutions. The application of ML for TSP has primarily focused on Supervised Learning (SL) and Reinforcement Learning (RL). However, SL methods encounter the challenge of expensive annotations, while RL methods struggle with sparse reward problems. 

Recently, \cite{min2023unsupervised} proposes a new approach named UTSP that employs Unsupervised Learning (UL) to build a data-driven heuristics for the TSP. This unsupervised method does not depend on any labelled dataset and generates a heatmap in a non-autoregressive manner, offering a distinct alternative to traditional SL and RL models. 

While the UL heuristics offer a promising approach, the challenge of generalizing across varying sizes and distributions remains significant. In particular, the model presented in \cite{min2023unsupervised} requires retraining to adapt to new sizes, indicating that a model trained on one size cannot effectively generalize to different sizes.

This paper explores the generalization capabilities of unsupervised heuristics for the TSP. Our findings indicate that the UL model is able to generalize across different problem sizes. Regarding the generalization behavior of different distributions, based on the hardness results by \cite{gent1996tsp}, we relate different distributions to distinct levels of hardnesses. This allows us to investigate the impact of the training data's hardness on the model's performance.

Our primary contributions are outlined as follows: We propose a novel approach for enabling a TSP model, once trained, to generalize effectively across different problem \textbf{sizes}. We show that training with larger problem sizes can enhance model performance. Furthermore, we investigate the impact of various \textbf{embedding dimensions} on TSP performance, finding that larger embedding dimensions can  build more effective representations to guide the search process. Additionally, we explore how the model performs when trained on datasets of varying \textbf{distributions}. Our findings indicate that models trained on harder instances exhibit better performance, which underscores the importance of training instances' distribution within the framework of UL for solving CO problems like the TSP.

While recent research papers explored using data-driven techniques for CO problems, most have focused on SL or RL. Very few have examined the generalization behaviours, particularly how training data (different distributions of TSP instances) influences final model performance~\cite{bi2022learning}. Our work addresses this gap, offering insights into the significance of training data selection and its direct impact on the effectiveness of ML models for CO tasks. This exploration contributes to understanding ML models in CO and provides practical guidelines for improving model generalization and performance in solving TSP.

\section{Related works}
\subsection{RL for TSP}
The goal of using RL for CO is to train an agent capable of either maximizing or minimizing the expected sum of future rewards, known as the return. For a given policy, the expected return from a current state is defined as the value function.  In the context of TSP, RL typically focuses on minimizing the length of the predicted route~\cite{ye2023deepaco,zhou2023towards,chen2023efficient,ma2023learning}. For example,
\cite{kool2018attention}  proposes a model based on attention layers and trains
the model using RL using a deterministic
greedy rollout. \cite{bello2016neural} trains a recurrent neural network to predict permutations of city coordinates and optimizes the parameters with a policy gradient method using the negative tour length as a reward signal.

However, as the size of the TSP increases, the rewards become increasingly sparse, necessitating long exploration steps before the agent achieves a positive return. So the RL setting is challenging as it only learns once the agent, randomly or through more sophisticated strategies, finds a better solution. Additionally, within RL, the learning process is hard to converge,  and the process may become trapped in local minima, as discussed in ~\cite{bengio2021machine}.

\subsection{SL For TSP}
In SL, the model is trained with a dataset including input coordinates alongside their corresponding optimal TSP solutions. The objective is to identify a function that predicts outputs for any given input coordinates, aiming for these predictions to approximate the optimal solutions~\cite{li2023t2tco,sun2023difusco,fu2021generalize}.
For example, \cite{xin2021neurolkh} trains a Sparse Graph Network using SL to evaluate edge scores, which are then integrated with the Lin-Kernighan-Helsgaun (LKH) algorithm to guide its search process. \cite{fu2021generalize} uses a 
GNN to learn from solved optimal solutions. The model is trained on a small-scale instances, which could be used to build larger heat maps.

However, In SL, the generation of optimal solutions for training is time-consuming. Finding optimal or near-optimal solutions for large TSP instances requires significant computational resources and sophisticated algorithms.  

In other words, an ideal model should circumvent these issues, avoiding the sparse reward problem in RL and not relying on labelled optimal solutuons in SL.  Addressing this, a recent approach by \cite{min2023unsupervised}  uses unsupervised learning (UL) and trains a GNN using a surrogate loss. The model generates heat maps through a non-autoregressive process, without relying on labelled optimal solutions or requiring the agents to explore better solutions, thereby circumventing the need for expensive annotation and mitigating the sparse reward problem.

This paper is structured as follows: Section~\ref{sec:UL4TSP} introduces the background of UL for TSP. Section~\ref{sec:sizeg} presents a method for generalizing across various problem sizes. Section~\ref{sec:exp} investigates the generalization behavior w.r.t. different embedding dimensions and training sizes. Finally, Section~\ref{sec:hardnessg} explores the generalization across different distributions through the lens of instance hardness.
\section{UL for TSP}
\label{sec:UL4TSP} 
Let's revisit the definition of the TSP. Essentially, the TSP can be reinterpreted as identifying the shortest Hamiltonian Cycle that encompasses all the cities.
In UL for TSP, the authors first reformulate the TSP into two constraints: the shortest path constraint and the Hamiltonian Cycle constraint. Subsequently, they construct a proxy for each of these constraints~\cite{min2023unsupervised}. 

In UTSP, given $n$ cities and their coordinates $(x_i,y_i) \in \mathbb{R}^2$, UTSP first uses GNN to generate a soft indicator matrix $\mathbb{T} \in \mathbb{R}^{n\times n}$ and use $\mathbb{T}$ to build the heat map $\mathcal{H} \in \mathbb{R}^{n \times n}$.   Row \(i\) of \(\mathcal{H}\) represents the probability distribution of directed edges originating from city \(i\), while column \(j\) corresponds to the probability distribution of directed edges terminating in city \(j\). This heat map is subsequently used to direct a local search algorithm. 
As mentioned, the goal of UTSP is to construct a proxy for two constraints. For the shortest constraint, the authors optimize the distance term: $ \langle \mathbf{D}, \mathcal{H} \rangle = \sum_{i=1}^n \sum_{j=1}^n \mathbf{D}_{i,j} \mathcal{H}_{i,j}$, where $\langle \cdot, \cdot \rangle$ is the Frobenius inner product, $\mathbf{D} \in \mathbb{R}^{n \times n}$ is the distance matrix and $\mathbf{D}_{ij}$ is the distance between city $i$ and city $j$. To address the Hamiltonian Cycle constraint, the authors introduce the $\mathbb{T} \rightarrow \mathcal{H}$ transformation, which is designed to implicitly encode this constraint.

\subsection{Understanding $\mathbb{T} \rightarrow \mathcal{H}$ transformation}
$\mathbb{T} \rightarrow \mathcal{H}$ transformation is defined as:
\begin{equation}\label{eq:tvt}
\mathcal{H} = \mathbb{T} \mathbb{V} \mathbb{T}^T,
\end{equation}
where $$\mathbb{V}  = 
\begin{pmatrix}
0 & 1 & 0 & 0 & \cdots & 0  & 0 & 0 \\
0 & 0 & 1 & 0 & \cdots & 0  & 0 & 0\\
0 & 0 & 0 & 1 & \cdots & 0  & 0 & 0\\

\vdots  & \vdots  & \vdots & \ddots  & \ddots & \vdots & \vdots  & \vdots  \\
0 & 0 & 0 & 0 & \ddots & 1 & 0 & 0\\
0 & 0 & 0 & 0 & \cdots & 0 & 1 & 0\\
0 & 0 & 0 & 0 & \cdots & 0 & 0 & 1\\
1 & 0 & 0 & 0 & \cdots & 0 & 0 & 0
\end{pmatrix}$$ is the shift matrix, where $\mathbb{V} \in \mathbb{R}^{n\times n}$. We can interpret $\mathbb{V}$ as representing a Hamiltonian cycle that follows the path $1\rightarrow 2 \rightarrow 3 \rightarrow \cdots \rightarrow n \rightarrow 1$, while $\mathbb{T}$ serves as an approximation of a general permutation matrix. Given that our initial heat map $\mathbb{V}$ represents a Hamiltonian cycle, and considering that both the Hamiltonian cycle constraint holds and the node ordering is equivariant under permutation operations, the Hamiltonian cycle constraint is implicitly encoded in this framework. For more details, we refer the reader to \cite{min2023unsupervised2}.

\begin{figure}[ht]
\vskip 0.2in
\begin{center}
\centerline{\includegraphics[width=\columnwidth]{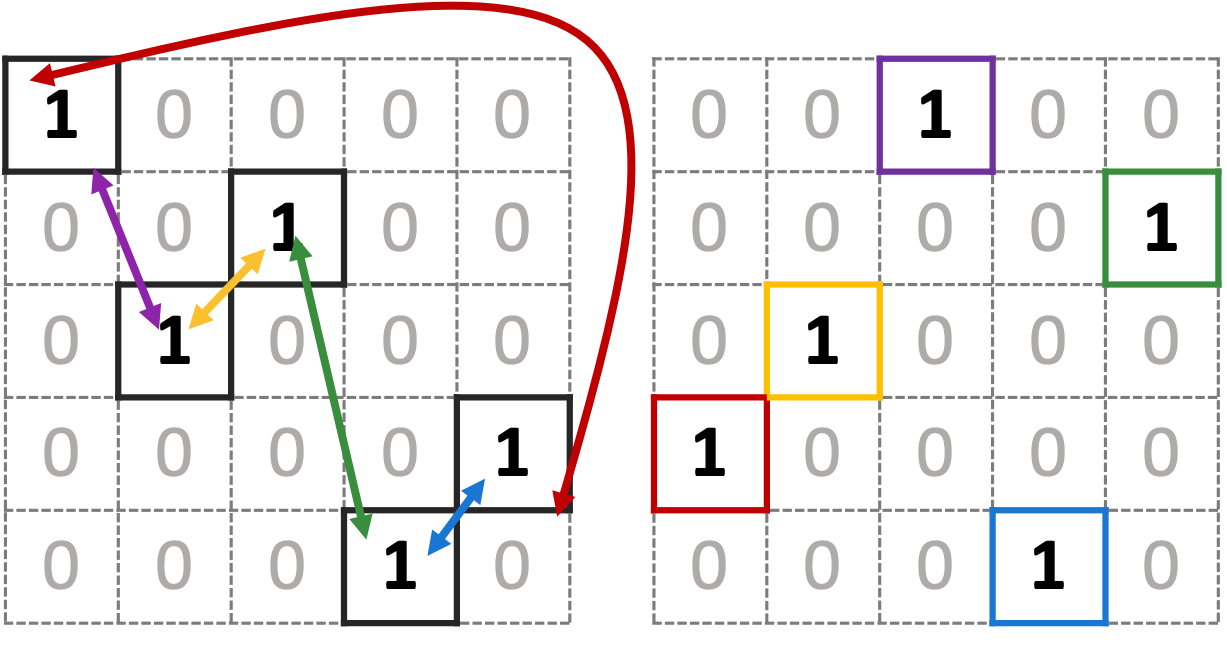}}
\caption{Illustration of $\mathbb{T}$ and the corresponding $\mathcal{H}$. $p_1$[1] = $p_2$[3] = $p_3$[2] = $p_4$[5] = $p_5$[4] = 1. This means there is a corresponding Hamiltonian Cycle: $1\rightarrow 3 \rightarrow 2 \rightarrow 5 \rightarrow 4 \rightarrow 1$.}
\label{fig:utsp}
\end{center}
\vskip -0.2in
\end{figure}

We can also write $\mathbb{T} \rightarrow \mathcal{H}$ transformation as:
\begin{equation}\label{eq:TH}
\mathcal{H} = \sum_{t=1}^{n-1}p_t p^T_{t+1} + p_np_1^T,
\end{equation}
where $p_t \in \mathbb{R}^{n \times 1}$ is the $t_{th}$ column of $\mathbb{T}$, $\mathbb{T} = [p_1|p_2|...|p_n]$.
Equation~\ref{eq:TH} provides another way of understanding the $\mathbb{T} \rightarrow \mathcal{H}$ transformation. The elements in $\mathcal{H}$ are defined using two nearest columns in $\mathbb{T}$. As shown in Figure~\ref{fig:utsp},  $p_1 = [1,0,0,0,0]^T$ and $p_2 = [0,0,1,0,0]^T$. Since the non-zero element in \(p_1\) is located at the first position and the non-zero element in \(p_2\) is at the third position, it indicates a directed edge from node 1 to node 3 in the heat map $\mathcal{H}$. This is depicted as the purple edge in Figure~\ref{fig:utsp}. Similarly, the presence of a non-zero element at the second position in $p_3$ implies that there is a directed edge from node 3 to node 2 in the heat map $\mathcal{H}$, represented by the yellow edge.

\subsection{Training UTSP}
In UTSP, the author train the model using the following loss $\mathcal{L}$ is: \begin{equation} \label{eq:loss}
\begin{aligned}
\lambda_1 \underbrace{\sum_{i=1}^n (\sum_{j=1}^n \mathbb{T}_{i,j} - 1)^2}_{\text{Row-wise constraint}}  +  \lambda_2    \underbrace{\sum_{i=1}^n \mathcal{H}_{i,i}}_{\text{No self-loops}}  +  \underbrace{\sum_{i=1}^n \sum_{j=1}^n \mathbf{D}_{i,j} \mathcal{H}_{i,j}}_{\text{Minimize the distance }}. 
\end{aligned}
\end{equation}
Here, the \emph{Row-wise constraint} encourages $\mathbb{T}$ to behave like a doubly stochastic matrix, thus serving as a soft relaxation of a permutation matrix~\cite{min2023unsupervised2}. The \emph{No self-loops} term discourages self loops in $\mathcal{H}$, where $\lambda_2$ is the distance of \emph{self-loop},  the \emph{Minimize the Distance} term acts as a proxy for minimizing the distance of a Hamiltonian Cycle.

Although UTSP offers a promising unsupervised way to learn the heat maps,  a notable limitation of the model is its lack of generalization. Specifically, a model trained on TSP instances with $n$ cities cannot be applied to other instances, such as instances with $n+1$ or $n-1$ cities. This limitation arises due to $\mathbb{T}$ having a fixed dimension of $\mathbb{R}^{n \times n}$. Consequently, the model's architecture is inherently tied to the size of the training instances, restricting its adaptability to TSP instances of varying city counts.

\section{Size Generalization}
\label{sec:sizeg}

Recall the understanding of $\mathbb{T}\rightarrow \mathcal{H}$ transformation in Equation~\ref{eq:TH}. We can interpret that the GNN generates a $n$-dimensional embedding for each city. In our generalized model,  given TSP instances with different sizes, for each node in these instances, the GNN outputs an embedding of dimension 
$m$. Following this, a Softmax activation function is applied to each column of the embedding matrix, resulting in the generation of 
$\mathbb{T} \in \mathbb{R}^{n\times m}$.

We then build $\mathcal{H}$ using\footnote{It is important to observe that when \(m \neq n\), \(\mathcal{H}\) is not doubly stochastic. We also tried either replacing \(\mathbb{T}\) with \(\sqrt{\frac{n}{m}}\mathbb{T}\) or substituting \(\mathcal{H}\) with \(\frac{n}{m}\mathcal{H}\), both of which yield similar outcomes.}:
\begin{equation}\label{eq:THg}
\mathcal{H} = \sum_{t=1}^{m-1}p_t p^T_{t+1} + p_mp_1^T,
\end{equation}
where $p_t \in \mathbb{R}^{n \times 1}$ is the $t_{th}$ column of $\mathbb{T}$.  Equation~\ref{eq:THg} can be reformulated analogously to Equation~\ref{eq:tvt} with $\mathbb{V} \in \mathbb{R}^{m \times m}$.

\begin{figure*}[ht]
\includegraphics[width=1\linewidth]{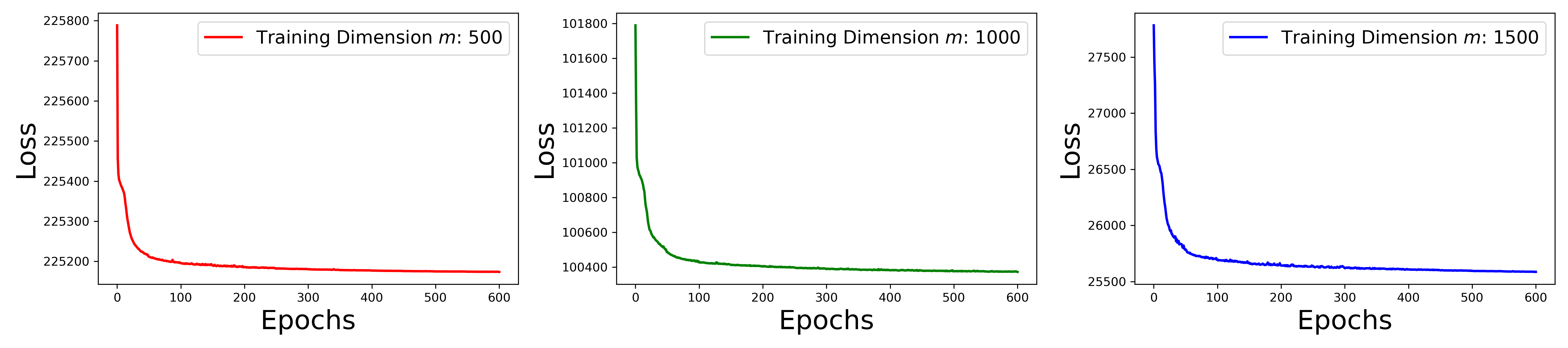}\caption{Training history of different $m$ (embedding dimension) on TSP-2000.} \label{fig:SizeGenS2000}
\end{figure*}
\begin{figure*}[ht]
\includegraphics[width=1\linewidth]{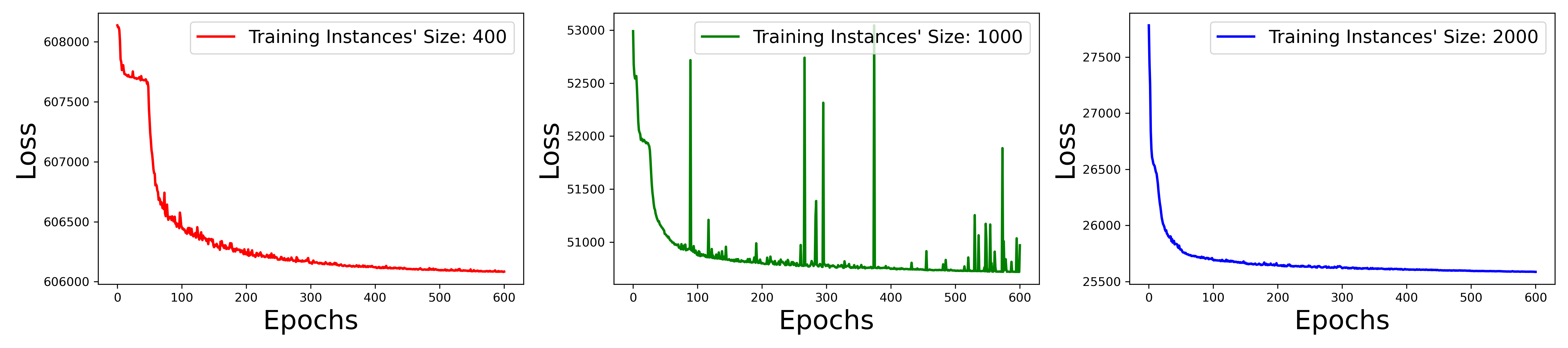}\caption{Training history of different $n$ (instance size) with same  embedding dimension $m=1500$.} \label{fig:SizeGenD1500}
\end{figure*}
In practice, we train our model under the loss $\mathcal{L}$:
\begin{equation} \label{eq:gloss}
\begin{aligned}
& \lambda_1 \underbrace{\bigl\{ \sum_{j=1}^n(1-\sum_{i=1}^n \mathcal{H}_{i,j})^2 + \sum_{i=1}^n(1-\sum_{j=1}^n \mathcal{H}_{i,j})^2 \bigl\} }_{\text{Row and column-wise constraint}} \\ &+  \underbrace{\sum_{i=1}^n \sum_{j=1}^n \mathbf{D}_{i,j} \mathcal{H}_{i,j}}_{\text{Minimize the distance }} . 
\end{aligned}
\end{equation}
By letting the GNN to output an $m$-dimensional embedding for each city, the model achieves generalization across different instances. This means that, through Equation~\ref{eq:TH}, the heat map $\mathcal{H}$ will consistently match the size of the input cities ($n \times n$ ).

\section{Experiment}
\label{sec:exp}
\begin{table*}[ht]
\setlength{\tabcolsep}{3pt} 
\footnotesize
\centering
\caption{Results of our generalizable model + Local Search w.r.t. existing baselines, tested on 128 instances with $n$ = 200, 500 and 1000.}
\vskip 0.15in
\label{table:table1}
    \begin{tabular}{lllllllllll}
    \toprule
       \multirow{2}{*}{Method} & \multirow{2}{*}{Type}  & \multicolumn{3}{c}{TSP200} & \multicolumn{3}{c}{TSP500} & \multicolumn{3}{c}{TSP1000}\\
         &  & Length & Gap (\%)  & Time & Length & Gap (\%)  & Time & Length & Gap (\%) & Time \\
        \hline
                Concorde &Solver & {10.7191} & {0.0000 } & {3.44m} & {16.5458} & {0.0000 } & {37.66m} & {23.1182} & {0.0000 } & {6.65h} \\
        Gurobi &Solver & {10.7036} & {-0.1446 } & {40.49m} & {16.5171} & {-0.1733 } & {45.63h} & {-} & {-} & {-}  \\
        LKH3 & Heuristic & {10.7195} & {0.0040 } & {2.01m} & {16.5463} & {0.0029 } & {11.41m} & {23.1190} & {0.0036 } & {38.09m}  \\
        GAT \citep{deudon2018learning} & RL, S & {13.1746} & {22.9079 } &  {4.84m} & {28.6291} & {73.0293 } &  {20.18m} & {50.3018} & {117.5860 } & {37.07m}  \\

        GAT \citep{kool2018attention}  & RL, BS & {11.3769} & {6.1364 } & {5.77m} & {19.5283} & {18.0257 } & {21.99m} & {29.9048} & {29.2359 } & {1.64h}  \\

        GCN \citep{joshi2019efficient} & SL, G & {17.0141} & {58.7272 } &  {59.11s} & {29.7173} & {79.6063 } & {6.67m} & {48.6151} & {110.2900 } &  {28.52m}  \\
        \hline
        \multirow{2}{*} {Att-GCRN\citep{fu2021generalize}} & \multirow{2}{*}{\tabincell{c}{SL+RL \\MCTS}} & \multirow{2}{*}{{10.7358}} & \multirow{2}{*}{{0.1563 }} & {20.62s} +  & \multirow{2}{*}{16.7471} & \multirow{2}{*}{{1.2169 }} & {31.17s} +  & \multirow{2}{*}{{23.5153}} & \multirow{2}{*}{{1.7179 }} & {43.94s} +   \\
        & & & & 1.33m & & & 3.33m & & & 6.68m\\ \hline
        \multirow{2}{*} {UTSP \citep{min2023unsupervised}} & \multirow{2}{*}{UL, Search} & \multirow{2}{*}{{10.7289}} & \multirow{2}{*}{{0.0918}} & {4.83s} +  & \multirow{2}{*}{16.6846} & \multirow{2}{*}{{0.8394}} & {7.28s} +  & \multirow{2}{*}{{23.3903}} & \multirow{2}{*}{{1.1770}} & {0.23m+}   \\
         & & & & 1.11m & & & 1.54m & & & 3.51m\\ \hline
        \multirow{2}{*} {Our Model} & \multirow{2}{*}{UL, Search} & \multirow{2}{*}{{10.7251}} & \multirow{2}{*}{\textbf{0.0558}} & {4.94s} +  & \multirow{2}{*}{16.6820} & \multirow{2}{*}{\textbf{0.8229}} & {5.66s} +  & \multirow{2}{*}{{23.3867}} & \multirow{2}{*}{\textbf{1.1616}} & {0.24m+}   \\
         & & & & 1.11m & & & 1.54m & & & 3.51m\\ 
\bottomrule
    \end{tabular}
\end{table*}
Here, we explore the impact of the generalized model on different problem sizes. Specifically, we study TSP  with 200, 500, and 1000 cities, each size is evaluated using 128 test instances. 

Different from previously UTSP setting,  our new methodology involves training models on larger datasets and testing them on smaller ones. Specifically, we train a model on a TSP-2000 dataset with $m=1500$ and test it on a TSP-1000 dataset; another model is trained on TSP-1000 with $m=800$ and tested on TSP-500; and finally, a model trained on TSP-400 with $m=320$ is tested on TSP-200. The TSP-2000, 1000, and 400 training datasets are created by randomly distributing points on a 2D plane, subject to a uniform distribution. For TSP-200 and TSP-400, we train the model for 300 epochs, while for TSP-1000, we train the model for 600 epochs. Each of these datasets consists of 5,000 training instances.

We train our model on one NVIDIA A100 Graphics Processing Unit, using the same Graph Neural Network (GNN) architecture as described in \cite{min2023unsupervised}. The model is trained on TSP instances of sizes 400, 1000, and 2000, using a configuration of two hidden layers, with each layer comprising 128 hidden units. The hyperparameter $\lambda_1$, as specified in Equation~\ref{eq:THg}, is set to 100.  Our test instances are taken from \cite{fu2021generalize}. Here, the performance gap is calculated using the  \(\frac{l - l_{opt}}{l_{opt}}\), where \(l\) represents the TSP length generated by our model and \(l_{opt}\) denotes the optimal length. We run the search algorithm on Intel Xeon Gold 6326.

In our approach, consistent with the existing UTSP framework, we employ the same search methodology. The process begins with the generation of the heat map $\mathcal{H}$, from which we extract the top $M$ largest values in each row. This extraction leads to the formation of a new heat map, denoted as $\Tilde{H}$. We  compute $\mathcal{H}' = \Tilde{H} + \Tilde{H}^T$ to symmetrize this updated heat map. $\mathcal{H}' $ is then used to guide the search process. 
We further calculate the overlap between non-zero edges in \(\mathcal{H}'\) and the optimal solutions, where a higher overlap ratio indicates that \(\mathcal{H}'\) more effectively covers the optimal solution. For more detailed information, we refer to ~\cite{min2023unsupervised}.

Our results are shown in Table~\ref{table:table1}, 
in the case of TSP-200, our model achieves a gap of \textbf{0.0558} \%, when tackling TSP-500, the model continues to demonstrate its robustness, with a gap of \textbf{0.8229}\%. The performance in both TSP-200 and TSP-500 suggests that our model's approach to guiding the local search is effective across various scales of the TSP.

When the model is applied to the largest tested instance size, TSP-1000, it achieves a gap of \textbf{1.1616}\%. which is the minimum one among all the methods. More importantly, it underscores the model's generalization to scale and maintain a level of efficiency in large-scale TSP instances. Our results across all three instance sizes illustrate that the model trained using Equation~\ref{eq:gloss} is able to generalize across instances of different sizes and  effectively enhances the search process.

\begin{figure*}[ht]
\setkeys{Gin}{width=1\linewidth}
\begin{minipage}[t]{0.24\textwidth}
\includegraphics{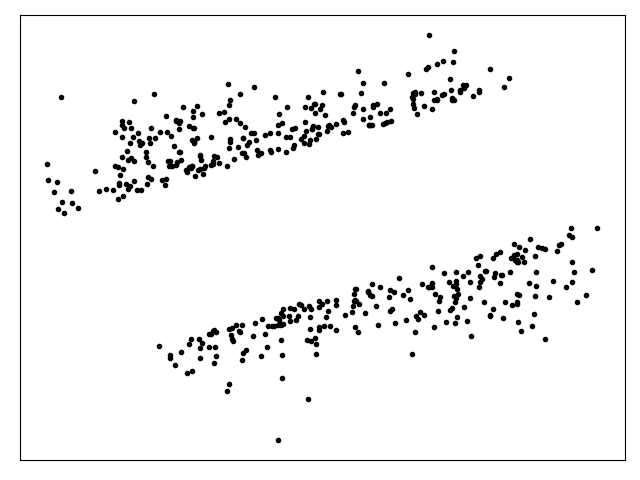}\caption{Expansion} \label{fig:Expansion}
\end{minipage}\hfill
\begin{minipage}[t]{0.24\textwidth}
\includegraphics{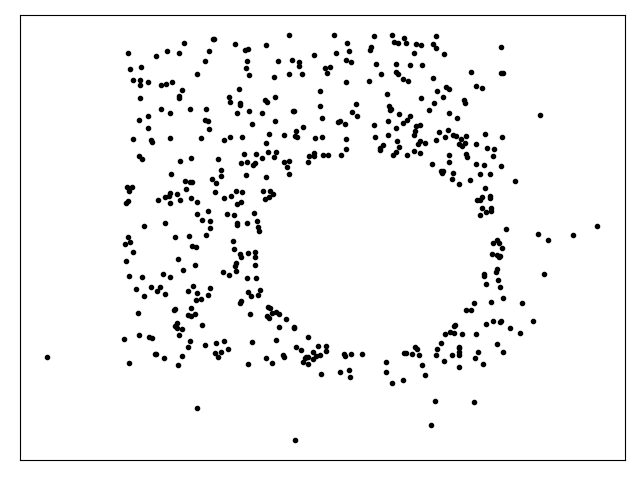}\caption{Explosion} \label{fig:Explosion}
\end{minipage}\hfill
\begin{minipage}[t]{0.24\textwidth}
\includegraphics{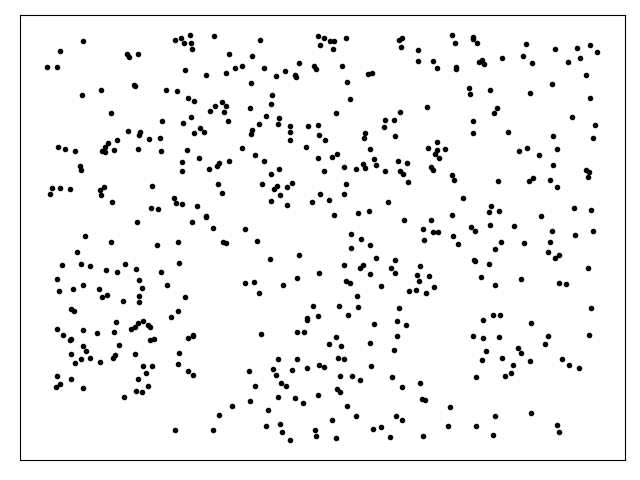}\caption{Implosion} \label{fig:Implosion}
\end{minipage}\hfill
\begin{minipage}[t]{0.24\textwidth}
\includegraphics{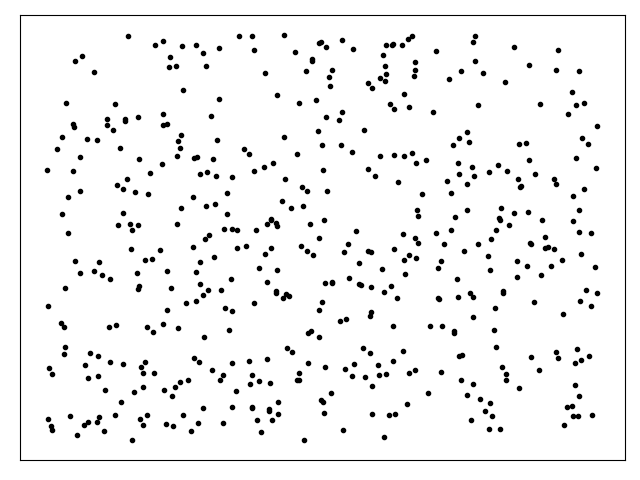}\caption{Uniform} \label{fig:Uniform}
\end{minipage}
\end{figure*}
\subsection{Impact of Varying $m$ on Training Performance}
As mentioned in Equation~\ref{eq:THg}, $m$ represents the embedding dimension of each node. In this study, we investigate the effect of the embedding dimension $m$ on the model's performance. Specifically, we train models on TSP-2000 instances with varying embedding dimensions: $m = 500, 1000,$ and $1500$. We then evaluate these models on TSP-1000 instances to assess their performance.
\begin{table}[H]
\setlength{\tabcolsep}{3pt} 
\caption{Overlap ratios and the search results on 128 TSP-1000 instances in \cite{fu2021generalize} using different embedding dimension $m$. We select top 5 elements from each row in the heat maps. }
\label{table:sizegen2000T5}
\vskip 0.15in
\begin{center}
\begin{small}
\begin{sc}
\begin{tabular}{lcccr}
\toprule
$m$& Overlap Ratio(\%)& Performance Gap(\%) \\
\midrule
500    & 82.70 & 2.0746 $\pm$ 0.5457 \\
1000 & 93.75 & 1.4832 $\pm$ 0.2305\\
1500    & 94.93  &  1.4145 $\pm$ 0.2005\\
\bottomrule
\end{tabular}
\end{sc}
\end{small}
\end{center}
\vskip -0.1in
\end{table}

The training curves for different embedding dimensions are shown in Figure~\ref{fig:SizeGenS2000}. 
We calculate the overlap ratios and search performance using models with different embedding dimensions, the results are shown in Table~\ref{table:sizegen2000T5},~\ref{table:sizegen2000T20}. Our findings indicate that an increase in the embedding dimension contributes to higher overlap ratios and enhanced search performance. For instance, the overlap ratio improves from 82.70\% to 94.93\% when the embedding dimension 
$m$ is increased from 500 to 1500, based on the heat maps with top 5 elements from each row. Correspondingly, the search performance also  improves, with the gap decreasing from 2.0746\% to 1.4145\%. This highlights the significance of embedding dimension in increasing model efficacy.
A larger embedding dimension can better identify optimal or near-optimal solutions and narrow the gap.
\begin{table}[H]
\setlength{\tabcolsep}{3pt} 
\caption{Overlap ratios and the search results on 128 TSP-1000 instances in \cite{fu2021generalize} using different embedding dimension $m$. We select top 20 elements from each row in the heat maps.}
\label{table:sizegen2000T20}
\vskip 0.15in
\begin{center}
\begin{small}
\begin{sc}
\begin{tabular}{lcccr}
\toprule
$m$& Overlap Ratio(\%)& Performance Gap(\%) \\
\midrule
500    &  99.99 & 1.1995 $\pm$ 0.1849\\
1000 & 100.00 & \textbf{1.1608} $\pm$ 0.1844 \\
1500    & 100.00 &  1.1616 $\pm$ 0.1743\\
\bottomrule
\end{tabular}
\end{sc}
\end{small}
\end{center}
\vskip -0.1in
\end{table}

Specifically, it is noteworthy that when selecting the top 20 elements from each row, both $m=1000$ and $m=1500$ achieve a 100.00\% overlap ratio, whereas $m=500$ does not cover all the optimal solutions, resulting in a larger gap. Furthermore, we observe that $m=1000$ exhibits marginally better performance compared to $m=1500$. This suggests that beyond a certain threshold, increasing the embedding dimension yields diminishing returns in terms of covering optimal solutions. It also implies that there might be an optimal range for the embedding dimension, indicating a need for careful consideration in the choice of $m$ to optimize model performance.

\subsection{Impact of Varying $n$ on Training Performance}
Our model can generalize across different sizes, meaning that training on one size can effectively translate to performance on another, previously unseen size. Here we investigate how varying the training size impacts the model's performance. We train the model using TSP-400, TSP-1000, and TSP-2000 instances, all with the same embedding dimension $m=1500$. The training results are illustrated in Figure~\ref{fig:SizeGenD1500}.

We then test how different training instances' sizes can affect the overlap ratio and the performance.  The results are shown in Table~\ref{table:sizegenD1500T5},~\ref{table:sizegenD1500T20}. We note that training with larger instances enhances search performance under both top 5 and top 20 conditions. Specifically, when selecting the top 5 elements from each row, the performance gap improves from 3.0762\% to 1.4145\%. Similarly, when choosing the top 20 elements from each row, the gap shows a marked improvement, decreasing from 1.1885\% to 1.1616\%.
\begin{table}[ht]
\setlength{\tabcolsep}{3pt} 
\caption{Overlap ratios and the search results on 128 TSP-1000 instances instances in \cite{fu2021generalize} with $m=1500$ using  training instances with different sizes. We select top 5 elements from each row in the heat maps. The first column denotes different training sizes.}
\label{table:sizegenD1500T5}
\vskip 0.15in
\begin{center}
\begin{small}
\begin{sc}
\begin{tabular}{lcccr}
\toprule
$n$& Overlap Ratio(\%)& Performance Gap(\%) \\
\midrule
400    & 68.83 & 3.0762 $\pm$ 1.3141 \\
1000 & 93.48 & 1.5563 $\pm$ 0.2345\\
2000    & 94.93  & 1.4145 $\pm$ 0.2005\\
\bottomrule
\end{tabular}
\end{sc}
\end{small}
\end{center}
\vskip -0.1in
\end{table}
\begin{table}[ht]
\setlength{\tabcolsep}{3pt} 
\caption{Overlap ratios and the search results on 128 TSP-1000 instances in \cite{fu2021generalize} with $m=1500$ using  training instances with different sizes.  We select top 20 elements from each row in the heat maps. The first column denotes different training sizes.}
\label{table:sizegenD1500T20}
\vskip 0.15in
\begin{center}
\begin{small}
\begin{sc}
\begin{tabular}{lcccr}
\toprule
$n$& Overlap Ratio(\%)& Performance Gap(\%) \\
\midrule
400    & 99.96 & 1.1885 $\pm$ 0.1927 \\
1000 & 100.00 & 1.1763 $\pm$ 0.1743\\
2000    & 100.00  & 1.1616 $\pm$ 0.1743\\
\bottomrule
\end{tabular}
\end{sc}
\end{small}
\end{center}
\vskip -0.1in
\end{table}
Our results highlight the importance of selecting larger training instance sizes to enhance model performance and efficiency.
\section{Hardness Generalization}
\label{sec:hardnessg}
Previous studies suggest that UL can generalize across different sizes, guide the search and reduce the search space,
Here, we delve into how UL's capability to reduce the search space is influenced by different distributions. Specifically, we explore the relationship between  different distributions and the efficiency of using UL for solving the TSP.

However, building a connection between various distributions and the efficacy of UL in reducing the search space presents significant challenges. To address this, we first focus on correlating different distributions with their hardness levels.

\paragraph{Phase transition}
A phase transition refers to a change in the solvability of NP-hard problems. When some parameters of the problem is varied, for example, the density of constraints in a Boolean 
satisfiability problem (SAT) problem~\cite{mitchell1992hard}, the problem undergoes a transition from being almost  solvable to unsolvable. To be specific,  The phase transition in SAT refers to a sharp change in the solvability of these problems, depending on the ratio of the number of clauses to the number of variables  in the formula. When the ratio is low (few clauses relative to variables), most instances of the problem are easy to solve. This is because there are fewer constraints, making it more likely to find a satisfying assignment.  Conversely, when this ratio is high (many clauses relative to variables), the problem becomes over-constrained, and most instances are also easy to solve because they are almost certainly unsatisfiable. The most interesting part occurs at a certain critical ratio, typically around 4.3 for 3-SAT problems. At this ratio, the problems undergo a phase transition and become extremely hard to solve. In other words, the problems are most difficult around the phase transition point
~\cite{monasson1999determining}.

Phase transitions provides a powerful framework to study the properties of NP-hard problems.
However, the exact nature and location of these transitions can be difficult to determine and may depend intricately on the structure of the different problems. For TSP, \cite{gent1996tsp} suggest using the parameter $\tau = l_{opt}/\sqrt{nA}$, where $A$ denotes the area covered by the TSP instance, $l_{opt}$ represents the length of the optimal solution, and $n$ is the number of cities. This approach is based on the observation that there is a rapid transition in solvability around a fixed value of the parameter, specifically at approximately $T_c = 0.78$.  
\begin{figure}[ht]
\vskip 0.2in
\begin{center}
\centerline{\includegraphics[width=\columnwidth]{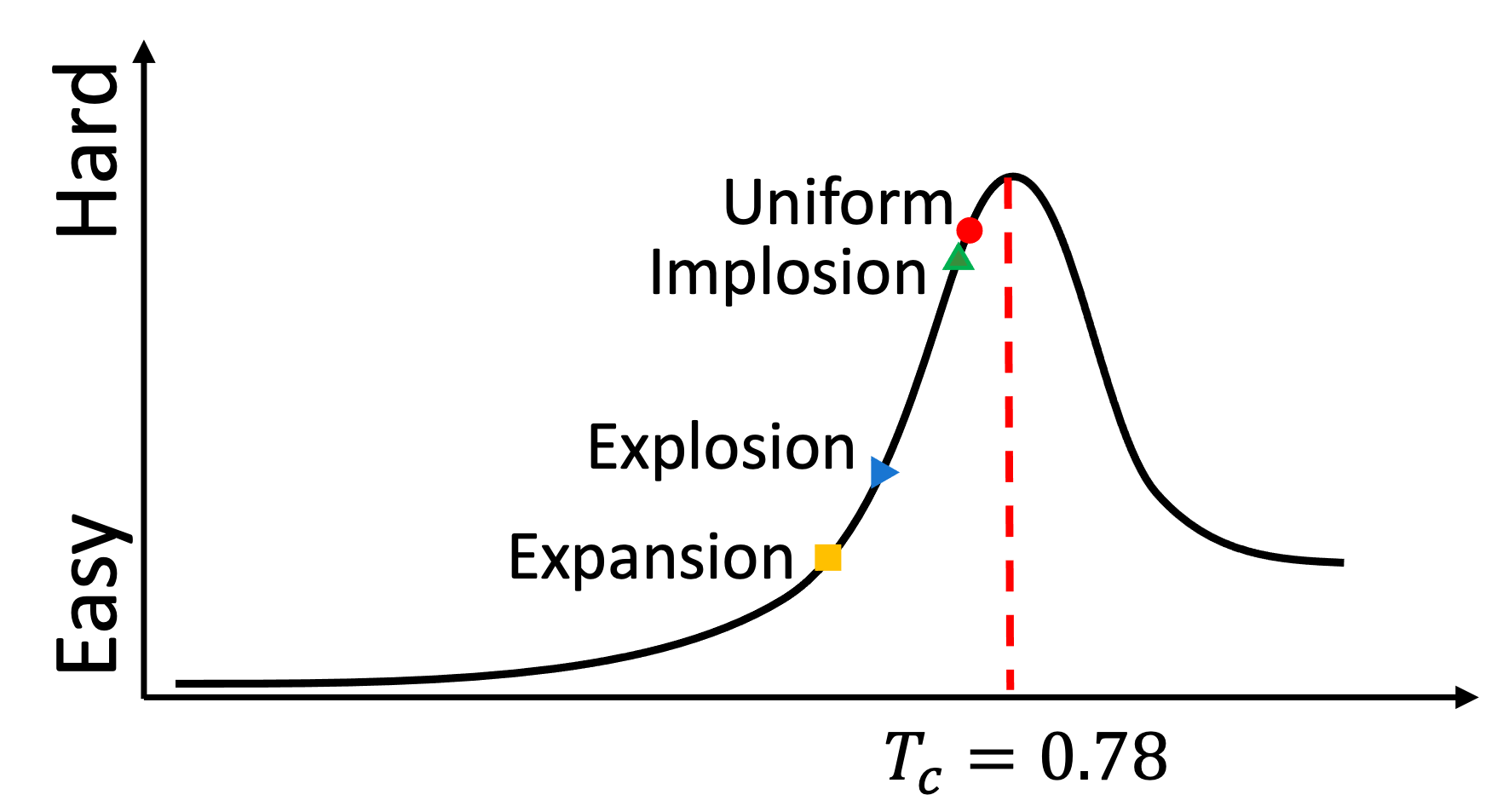}}
\caption{TSP phase transition and the $\tau$ values for different distributions.}
\label{fig:tsphardness}
\end{center}
\vskip -0.2in
\end{figure}

Here we study four different distributions and see how it can effect the search space reduction, an illustration of these four distribution is shown in Figure~\ref{fig:Expansion} $\sim$~\ref{fig:Uniform}. 
As mentioned earlier, around the phase transition point, the problems often exhibits the greatest computational complexity (Hard). Figure~\ref{fig:tsphardness} illustrates the scheme of phase transition in the TSP. The $x$-axis is the $\tau$ value, while the $y$-axis corresponds to the level of hardness. The point at which $\tau$ equals the critical threshold $T_c = 0.78$ marks the peak of difficulty, exhibiting the highest hardness, we refer more details to  ~\cite{gent1996tsp}.

Furthermore, we present the $\tau$ values for four different distributions, where each $\tau$ is computed as an average from 100 instances, each with a size of 200, 500 and 1000, detailed in Table~\ref{table:hardness} and Figure~\ref{fig:tsphardness}.
\begin{table}[ht]
\setlength{\tabcolsep}{3pt} 
\caption{$\tau = l_{opt}/\sqrt{nA}$ of Expansion, Explosion, Implosion, and Uniform for different sizes.}
\label{table:hardness}
\vskip 0.15in
\begin{center}
\begin{small}
\begin{sc}
\begin{tabular}{lcccr}
\toprule
Size &  Expansion &  Explosion &  Implosion &  Uniform \\
\midrule
1000 &     0.4838 &     0.5629 &     0.7237 &   0.7460 \\
500 &     0.5114 &     0.5905 &     0.7338 &   0.7515 \\
200 &     0.5796 &     0.6337 &     0.7539 &   0.7745 \\
\bottomrule
\end{tabular}
\end{sc}
\end{small}
\end{center}
\vskip -0.1in
\end{table}

As shown in Figure~\ref{fig:tsphardness}, the Uniform distribution is closest to the phase transition point $T_c$. This  indicates a highest level of hardness. Consequently, in terms of transitioning from hard to easy, the order is observed as follows: Uniform $\approx$ Implosion $>$ Explosion $>$  Expansion. Following upon this concept, we examine how these distributions influence the capacity of UL to efficiently  reduce the search space and guide the search.
\begin{figure}[ht]
  \centering
    \includegraphics[width=\columnwidth]{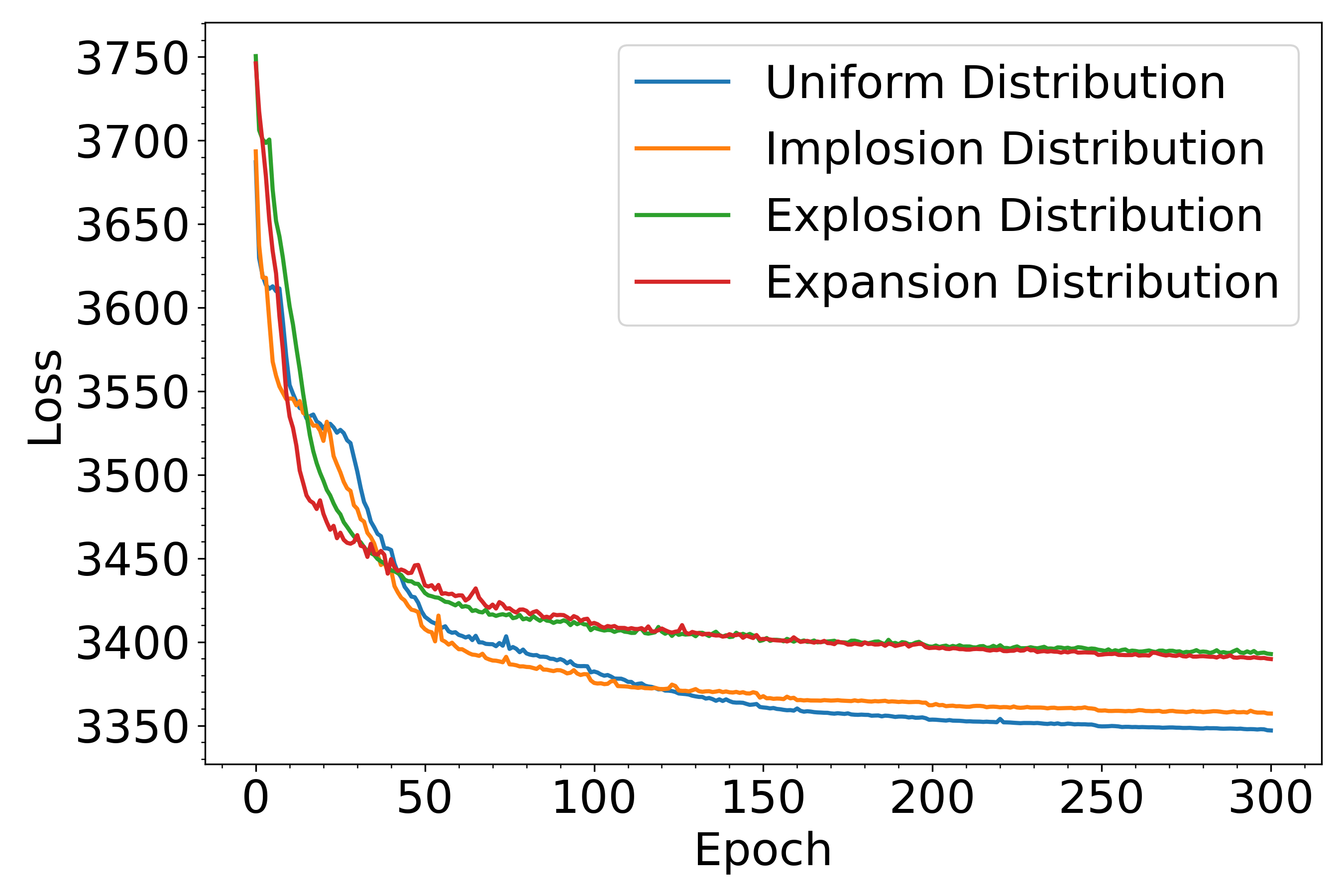}
    \caption{The training curves for TSP-400 with $m=320$ across four different distributions are shown; the model is then tested on 128 TSP-200 instances.}
    \label{fig:train200}
\end{figure}
\begin{figure}[ht]
    \includegraphics[width=\columnwidth]{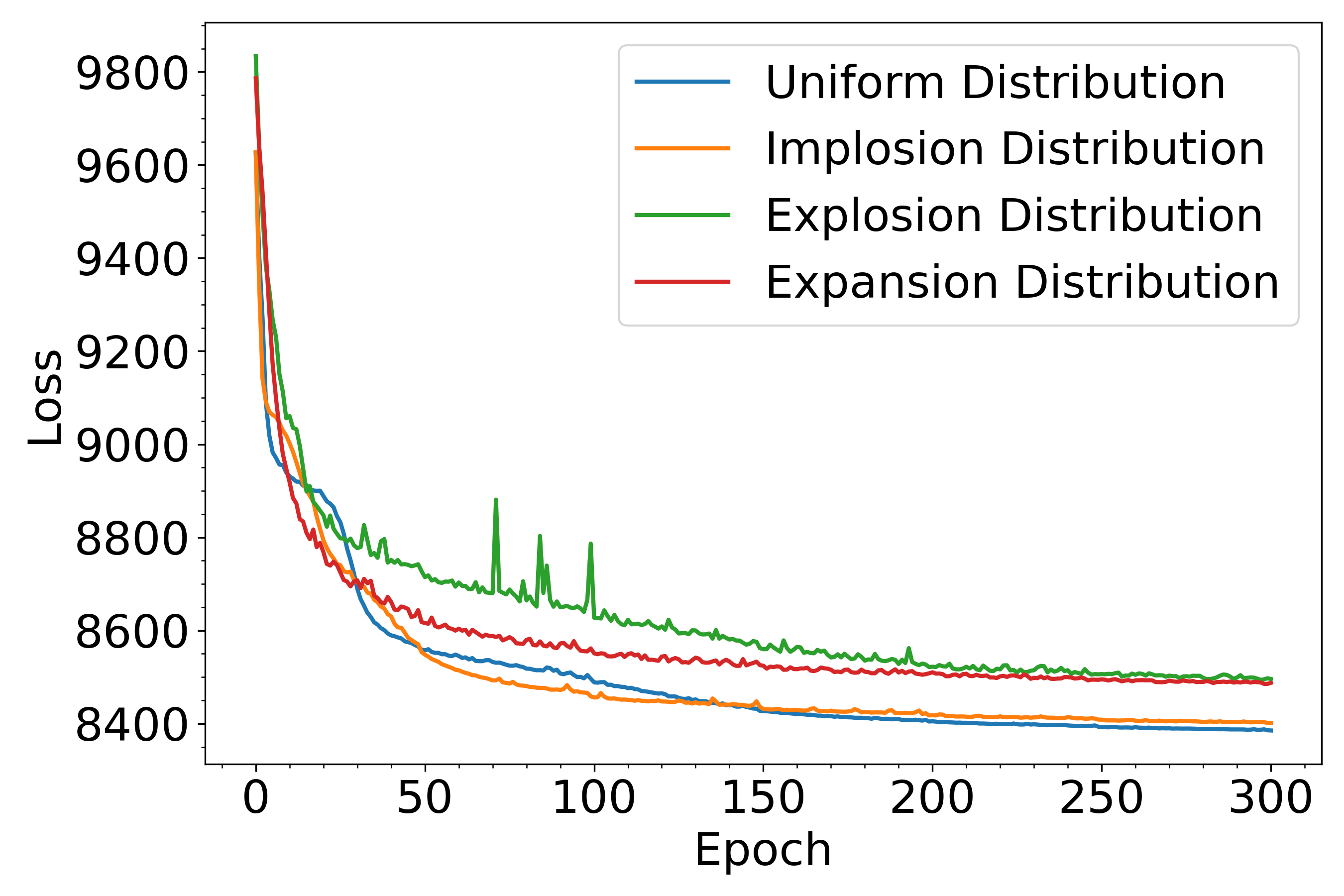}
    \caption{The training curves for TSP-1000 with $m=800$ across four different distributions are shown; the model is then tested on 128 TSP-500 instances.}
    \label{fig:train500}
\end{figure}
\begin{table}[ht]
\setlength{\tabcolsep}{3pt} 
\caption{Overlap ratios and the search results on 128 TSP-200 instances in \cite{fu2021generalize} using different distributions. We select top 5 elements from each row in the heat maps.}
\label{table:tsp200}
\vskip 0.15in
\begin{center}
\begin{small}
\begin{sc}
\begin{tabular}{lcccr}
\toprule
Dataset & Overlap Ratio(\%)& Performance Gap(\%) \\
\midrule
Uniform    & 95.64& 0.0883$\pm$ 0.0885 \\
Implosion & 95.50& 0.0876$\pm$ 0.0920\\
Explosion    & 95.29& 0.0979$\pm$ 0.0907\\
Expansion    & 94.00 & 0.1131$\pm$ 0.0973 \\
\bottomrule
\end{tabular}
\end{sc}
\end{small}
\end{center}
\vskip -0.1in
\end{table}
\begin{table}[ht]
\setlength{\tabcolsep}{3pt} 
\caption{Overlap ratios and the search results on 128 TSP-500 instances in \cite{fu2021generalize} using different distributions. We select top 5 elements from each row in the heat maps.}
\label{table:tsp500}
\vskip 0.15in
\begin{center}
\begin{small}
\begin{sc}
\begin{tabular}{lcccr}
\toprule
Dataset & Overlap Ratio(\%)& Performance Gap(\%) \\
\midrule
Uniform    & 95.47& 0.9311$\pm$ 0.1638 \\
Implosion & 95.40& 0.9394$\pm$ 0.1732\\
Explosion    & 94.99& 0.9410$\pm$ 0.1764\\
Expansion    & 94.03& 1.0137$\pm$ 0.1800\\
\bottomrule
\end{tabular}
\end{sc}
\end{small}
\end{center}
\vskip -0.1in
\end{table}
\begin{table}[ht]
\setlength{\tabcolsep}{3pt} 
\caption{Overlap ratios and the search results on 128 TSP-1000 instances in \cite{fu2021generalize} using different distributions. We select top 5 elements from each row in the heat maps.}
\label{table:tsp1000}
\vskip 0.15in
\begin{center}
\begin{small}
\begin{sc}
\begin{tabular}{lcccr}
\toprule
Dataset & Overlap Ratio(\%)& Performance Gap(\%) \\
\midrule
Uniform    & 94.93 &  1.4145 $\pm$ 0.2005 \\
Implosion & 94.71 & 1.4060 $\pm$ 0.2078\\
Explosion    & 93.86  & 1.5274 $\pm$ 0.2632\\
Expansion    & 93.38 & 1.5777 $\pm$ 0.2735\\
\bottomrule
\end{tabular}
\end{sc}
\end{small}
\end{center}
\vskip -0.1in
\end{table}
\begin{figure*}[ht]
\vskip 0.2in
\begin{center}
\centerline{\includegraphics[width=\textwidth]{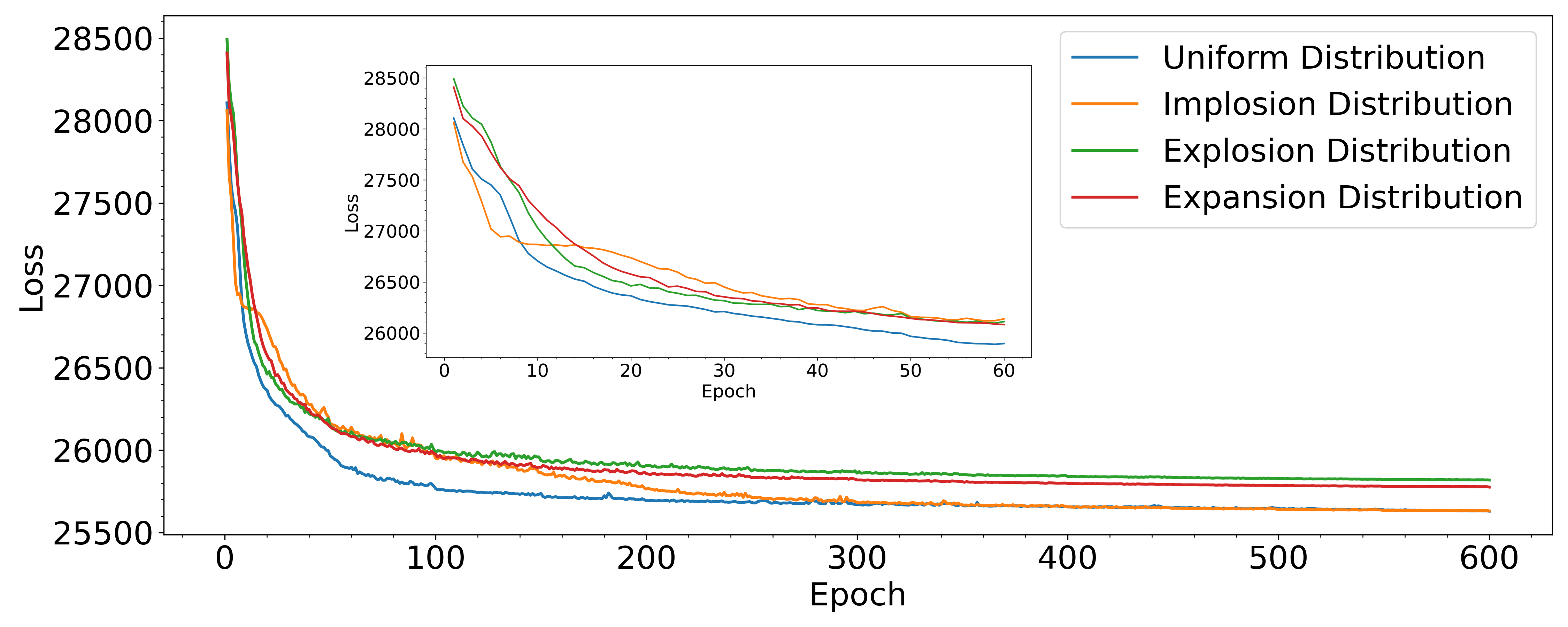}}
\caption{The training curves on TSP-2000 with $m=1500$ across four different distributions are shown; the model is then tested on 128 TSP-1000 instances.}
\label{fig:LosswTraining}
\end{center}
\vskip -0.2in
\end{figure*}

We first train the models using 4 different distributions with the same parameters in Section~\ref{sec:sizeg}. We calculate the overlap ratio of these models for TSP-200, 500, and 1000. The training results are shown in Figure~\ref{fig:train200},~\ref{fig:train500} and \ref{fig:LosswTraining}.  We observe that models trained with harder instances consistently exhibit a lower loss. Specifically, the loss curves for models trained using the Uniform distribution consistently show the lowest loss, while those trained with Expansion and Explosion distributions demonstrate higher losses.  
This suggests that the hardness level of training instances plays a significant role in the effectiveness of the model training, directly impacting the loss metrics. It is important to note that throughout our training process, all other hyperparameter settings remained constant. Therefore, the observed variations in loss can be attributed solely to the differences in training distributions.  
\begin{table}[ht]
\setlength{\tabcolsep}{3pt} 
\caption{Overlap ratios and the search results on 128 TSP-1000 instances in \cite{fu2021generalize} using different distributions. We select top 20 elements from each row in the heat maps.}
\label{table:tsp1000t20}
\vskip 0.15in
\begin{center}
\begin{small}
\begin{sc}
\begin{tabular}{lcccr}
\toprule
Dataset & Overlap Ratio(\%)& Performance Gap(\%) \\
\midrule
Uniform    & 100.00 & \textbf{1.1616}$\pm$ 0.1743 \\
Implosion & 100.00 & 1.1844$\pm$ 0.1572\\
Explosion    & 100.00  & 1.1937$\pm$ 0.1764\\
Expansion    & 100.00& 1.1797$\pm$  0.2025\\
\bottomrule
\end{tabular}
\end{sc}
\end{small}
\end{center}
\vskip -0.1in
\end{table}

We then evaluate how different distributions can affect the search results. We pick the top 5 element each row and build the heat maps. The overlap ratio and the search results are shown in Table~\ref{table:tsp200},~\ref{table:tsp500} and \ref{table:tsp1000}. When training on easier distributions such as Explosion and Expansion, we observe low overlap ratios and larger performance gaps. This indicates that models trained on simpler distributions may struggle to generalize effectively to more challenging instances of the problem. The lower overlap ratios suggest that the solutions generated by these models are less aligned with the optimal solutions, and the larger performance gaps highlight a significant disparity in effectiveness when these models are applied to the test TSP instances. Training on harder distributions, such as Uniform, yields higher overlap ratios and improved search performance. This  indicates that models trained on harder distributions can build a better representation of the search space, which enables the search to perform more effectively. It is also observed that the \emph{plateaus} during training are more pronounced when training on harder instances, suggesting that the optimization landscape becomes more complex when the hardness level increases.

We evaluate the model's performance on TSP-1000 instances by utilizing the top 20 elements from each row for each distribution, as detailed in Table~\ref{table:tsp1000t20}. We observe that by selecting the top 20 elements, \(\mathcal{H}'\) is able to cover 100.00\% of the optimal solutions. Overall, the performance gaps across the distributions are similar, with training on uniform distributions continuing to exhibit the lowest performance gap.

\section{Conclusion}
This work introduces a new methodology that allows a trained, unsupervised TSP model to generalize across different problem sizes. Our results demonstrate that training on larger problem instances can improve performance compared to training with smaller instances. Additionally, we delve into the influence of embedding dimensions on TSP results, showing that larger embedding dimensions are important in constructing more effective representations that guide the search process more efficiently. Moreover, we investigate the model's performance using training datasets with different levels of hardnesses. We show that training on harder instances can improve model performance, emphasizing the importance of selecting training instances with appropriate difficulty levels. We train our models on different TSP distributions to understand their impact on the effectiveness of UL models. 
Our study indicates a clear relationship between the inherent hardness of distribution and the model's capacity to generalize and effectively solve TSP instances. To our knowledge, this is the first study to systematically investigate and demonstrate this connection.

Our results highlight the  relationship between the characteristics of training instances (size and hardness), embedding dimensions, and model performance in UL, particularly when addressing CO problems such as the TSP. We anticipate that these findings — emphasizing the benefits of training on larger, harder instances with increased embedding dimensions — can inspire further research in the application of Unsupervised Learning to Combinatorial Optimization tasks.

\section{Acknowledgement}
This project is partially supported by the Eric and Wendy Schmidt AI
in Science Postdoctoral Fellowship, a Schmidt Futures program; the National Science Foundation
(NSF) and the  National Institute of Food and Agriculture (NIFA); the Air
Force Office of Scientific Research (AFOSR);  the Department of Energy;  and the Toyota Research Institute (TRI).


\bibliography{icml2024}
\bibliographystyle{icml2024}




\end{document}